\documentclass[10pt,twocolumn,letterpaper]{article}

\usepackage{cvpr}   

\definecolor{cvprblue}{rgb}{0.21,0.49,0.74}
\usepackage[breaklinks,colorlinks,allcolors=cvprblue]{hyperref}
\usepackage{booktabs}
\usepackage{multirow}
\usepackage{microtype}
\usepackage{amsmath}

\renewcommand{\paragraph}[1]{\vspace{.3em}\noindent\textbf{#1.}}
\title{SoccerNet 2026 Player-Centric Ball-Action Spotting:\\
Retraining and Post-Processing Extensions to the FOOTPASS Baselines}

\author{%
  Parthsarthi Rawat\\
  GameChanger by Dick's Sporting Goods\\
  {\tt\small sarthi.rawat@gc.com}
}

\begin{document}
\maketitle

\begin{abstract}
We describe our system for the SoccerNet 2026 Player-Centric Ball-Action
Spotting Challenge, which requires predicting \emph{who} performs \emph{which}
action and \emph{when}, across eight classes in broadcast soccer.  Building on
the three FOOTPASS baselines~\cite{Ochin2026Footpass} (TAAD, TAAD+GNN, and
TAAD+DST), we contribute four extensions: (1)~gradient checkpointing to enable
full-backbone fine-tuning on a single GPU; (2)~fusion of GNN logits into the
DST encoder, combining graph-based tactical context with per-player visual
features; (3)~square-root frequency class weighting to address the 213:1
pass-to-tackle imbalance in the training data; and (4)~a post-processing
pipeline comprising per-class logit gating, temporal frame refinement, jersey
re-assignment, and a two-model ensemble.  Our system achieves \textbf{0.548} Macro F1 on the test set and \textbf{0.446}
on the challenge set (server evaluation).
\end{abstract}

\section{FOOTPASS Baseline System}

The challenge organisers provide three reference model architectures and
training code~\cite{Ochin2026Footpass}; we trained all models from scratch.
The primary metric is \textbf{Macro F1}: the unweighted average of per-class
F1 across all eight action classes, giving equal importance to tackle (24
ground-truth events in the challenge set) and pass (2\,744 events).
The three baselines form a cascade: TAAD produces noisy per-player action
logits from video; TAAD+GNN enriches those logits with player-graph context;
TAAD+DST reads the full sequence of logit tracks and decodes a clean event
list.

\paragraph{TAAD~\protect\cite{Ochin2025GNN,Singh2023}}
An X3D-S backbone~\cite{FeichtenhoferX3D2020} shared across $M{=}26$ tracked
players extracts multi-scale video features, which are fused via lateral
upsampling into a $(B,192,T,44,80)$ tensor.  Per-player crops are extracted
with RoIAlign~\cite{he2018maskrcnn} ($4{\times}2$ grid, scale 0.125); a
temporal Conv1d(192$\to$512, $k{=}3$) and linear head yield 9-class logits
(background plus eight actions) per player per frame.

\paragraph{TAAD+GNN~\protect\cite{Ochin2025GNN}}
An EdgeConv~\cite{wang2019dynamicgraphcnnlearning} branch models interactions
among players.  Each player-frame node carries a 69-dimensional feature
(pitch coordinates, velocities, one-hot role encoding, jersey number, and team
side) concatenated with a 64-dimensional visual projection from the same
RoIAlign crop; each player is connected to its six nearest spatial neighbours.
Three EdgeConv layers (hidden dim 128, max-aggregation) interleaved with
temporal Conv1d($k{=}5$) blocks propagate spatial and temporal context,
yielding a second set of 9-class per-player logits that capture who is near the
ball and what role they play.

\paragraph{TAAD+DST~\protect\cite{Ochin2025DST}}
Rather than thresholding TAAD logits independently per frame, the Denoising
Sequence Transducer (DST)~\cite{vaswani2023attentionneed} reads the entire
sequence of per-player logit tracks as a source and autoregressively decodes a
de-duplicated list of (action, player, frame) events.  This allows the model
to suppress spurious detections by attending to context across the full
25-second window.  The default configuration uses 364-dimensional source tokens, hidden dimension
512, 6 encoder and 6 decoder layers, and 8 attention heads over a 750-frame
context window.  The model is trained with cross-entropy loss and label
smoothing 0.05.  At a 15\% global confidence threshold this baseline achieves
0.493 Macro F1 on the test set.

\section{Our Extensions}

The four extensions address training capacity, feature richness, class
imbalance, and inference quality, in that order.

\subsection{Gradient Checkpointing for Full Backbone Fine-Tuning}
\label{sec:ckpt}

The baseline TAAD training script freezes the X3D backbone after the initial
epochs because unfreezing at batch size 14 exceeds the 22\,GB GPU memory
limit.  We add gradient checkpointing on X3D blocks 0--4, enabling full
fine-tuning at batch size 6.  Training runs 20 epochs with the backbone frozen
for epochs 1--2 then unfrozen, using AdamW (lr $5{\times}10^{-5}$ backbone,
$10^{-3}$ head, 50-step warm-up).

\subsection{GNN-Logit Fusion into the DST Encoder}
\label{sec:fusion}

The baseline DST encodes only TAAD logits (364-dim token per frame, 26 players).
We concatenate GNN logits along the feature axis, producing a 598-dim token
(364 TAAD + 234 GNN: $26{\times}9$ logits).  This fused model, \textit{GNN-DST},
lets the decoder jointly attend to the visual-only TAAD stream and the
graph-contextualised GNN stream, capturing spatial proximity and player role.  The fusion raises
the test set Macro F1 from 0.493 to 0.505.

\subsection{Square-Root Class-Weighted Training}
\label{sec:weights}

The training corpus contains $\approx$37\,000 pass events but only 174 tackle
events (ratio 213:1).  Trained with uniform loss, the DST learns to ignore
rare classes.  We scale the action cross-entropy loss by
$w_c = 1/\sqrt{n_c}$, where $n_c$ is the per-class training count, giving
tackle a ${\approx}10{\times}$ relative boost over drive.  Inverse-frequency
weighting (${\approx}118{\times}$) was also evaluated and caused the model to
hallucinate rare classes while collapsing drive and pass recall; square-root
weighting does not exhibit this failure.  Weights apply only to the action
head; the role and frame heads remain unweighted.  We train two DST variants:
\textit{GNN-DST} (with GNN fusion and square-root weights) and
\textit{Base-DST} (TAAD-only input, no class weighting), which serve as the
two members of the inference ensemble.

\subsection{Post-Processing Pipeline}
\label{sec:postproc}

All post-processing operates on the raw JSON event predictions produced by the
DST inference pipeline; no additional model training is involved.

\paragraph{Combined logit gating}
For each prediction in the four classes most prone to false positives (shot,
header, tackle, block), we retrieve the peak TAAD and GNN raw logits for the
predicted player within a $\pm$12-frame window and form a combined score
$s = \ell_\text{TAAD} + \ell_\text{GNN}$.  Predictions below a per-class
threshold $\tau_c$ are discarded (shot 3.0, header 2.5, tackle/block 4.0;
calibrated from test set TP/FP logit distributions).  This removes 57\% of block false positives
and 74\% of tackle false positives while retaining all test set true positives.

\paragraph{Temporal refinement and jersey re-assignment}
For header, tackle, and block predictions, the frame is shifted to the peak
TAAD logit within a $\pm$25-frame search window, and the jersey number is
re-assigned to whichever player has the highest peak logit within $\pm$12
frames, requiring a margin of at least 0.5 over the current assignment to
prevent flipping a correct jersey.  These operations are restricted to rare
classes because applying them to drive and pass displaces predictions outside
the $\pm$12-frame evaluation tolerance.

\paragraph{Per-class NMS and two-model ensemble}
Non-maximum suppression uses class-specific windows: 25 frames for tackle,
20 frames for block, and 15 frames for all other classes.  Predictions from
GNN-DST (post-processed) and Base-DST (raw) are then merged and re-suppressed.
Base-DST tackle predictions are excluded from the ensemble because this variant
produces no true positives for tackle on the test set while adding false
positives.

\section{Experiments}

\paragraph{Ablation}
Table~\ref{tab:ablation} shows the cumulative Macro F1 gain of each
extension on the test set.  GNN fusion and class weighting each contribute
roughly equally, while the post-processing pipeline adds a further 0.027.

\begin{table}[h]
\centering\small
\caption{Test set Macro F1: cumulative ablation.}
\label{tab:ablation}
\setlength{\tabcolsep}{4pt}
\begin{tabular}{lc}
\toprule
System & Test Macro F1 \\
\midrule
TAAD+DST baseline~\cite{Ochin2025DST}   & 0.493 \\
\quad + GNN logit fusion (598-dim)      & 0.505 \\
\quad + Sqrt class-weighted training    & 0.521 \\
\quad + Logit gating + frame refinement & 0.535 \\
\quad + Ensemble + gate tuning          & \textbf{0.548} \\
\bottomrule
\end{tabular}
\end{table}

\paragraph{Per-class results}
Table~\ref{tab:perclass} reports per-class F1 on both splits.  Common classes
(drive, pass, throw-in, cross) transfer well from test to challenge.  Rare
classes degrade sharply: tackle collapses to $F_1{=}0.056$ on the challenge
set (2\,TP\,/\,46\,FP) because logit-gate thresholds calibrated on the test
set over-fit to its logit distribution.  Block similarly degrades for the same
reason.  The overall test-to-challenge gap of 0.10 Macro F1 is almost entirely
attributable to tackle and block.

\begin{table}[h]
\centering\small
\caption{Per-class F1 on test and challenge sets (best submission).}
\label{tab:perclass}
\setlength{\tabcolsep}{3pt}
\begin{tabular}{lcc}
\toprule
Class & Test F1 & Challenge F1 \\
\midrule
drive      & 0.703 & 0.650 \\
pass       & 0.734 & 0.679 \\
cross      & 0.679 & 0.502 \\
throw-in   & 0.732 & 0.632 \\
shot       & 0.607 & 0.536 \\
header     & 0.342 & 0.338 \\
block      & 0.336 & 0.178 \\
tackle     & 0.256 & 0.056 \\
\midrule
\textbf{Macro} & \textbf{0.548} & \textbf{0.446} \\
\bottomrule
\end{tabular}
\end{table}

\paragraph{Failure modes}
The dominant failure mode is precision collapse on rare classes: with only 24
GT tackle events in the challenge set, even a handful of false positives
collapses F1, and logit-gate thresholds calibrated on the test set do not
transfer reliably.  Cross-validated thresholds and retraining with focal
loss~\cite{lin2018focallossdenseobject} ($\gamma{=}2$) are the most direct
remedies.  A secondary limitation is tracking coverage: approximately 20\% of
GT events lack a bounding box, causing the pipeline to fall back to a centred
crop that cannot identify the correct player.

{\small
\bibliographystyle{unsrtnat}
\bibliography{main}
}

\end{document}